\documentclass[journal,twoside,web]{IEEEtran}
\usepackage{cite}
\usepackage{amsmath,amssymb,amsfonts}
\usepackage{algorithmic}
\usepackage{graphicx}
\usepackage{textcomp}
\usepackage{wrapfig}
\usepackage{orcidlink}
\usepackage{booktabs}
\usepackage{multirow}

\definecolor{abstractbg}{rgb}{0.89804,0.94510,0.83137}
\begin{document}
\title{P2P-Insole: Human Pose Estimation Using Foot Pressure Distribution and Motion Sensors}
\author{Atsuya Watanabe, Ratna Aisuwarya\,\orcidlink{0000-0001-8337-9017},~\IEEEmembership{Student Member,~IEEE}, Lei Jing\orcidlink{0000-0002-1181-2536},~\IEEEmembership{Member,~IEEE}
\thanks{Atsuya Watanabe is with the Department of Computer Science and Engineering, University of Aizu, Aizu-Wakamatsu, Fukushima, Japan (e-mail: wenyed56@gmail.com)}
\thanks{Ratna Aisuwarya is with the Department of Computer Engineering, Andalas University, Padang, West Sumatra, Indonesia, and also with the Department of Computer Science and Engineering, University of Aizu, Aizu-Wakamatsu, Fukushima, Japan (e-mail: aisuwarya@it.unand.ac.id)}
\thanks{Corresponding author: Lei Jing, Professor of the Department of Computer Science and Engineering, University of Aizu, Aizu-Wakamatsu, Fukushima, Japan (e-mail: leijing@u-aizu.ac.jp)}}

\IEEEtitleabstractindextext{%
\fcolorbox{abstractbg}{abstractbg}{%
\begin{minipage}{\textwidth}%
\begin{wrapfigure}[12]{r}{3in}%
\includegraphics[width=3in]{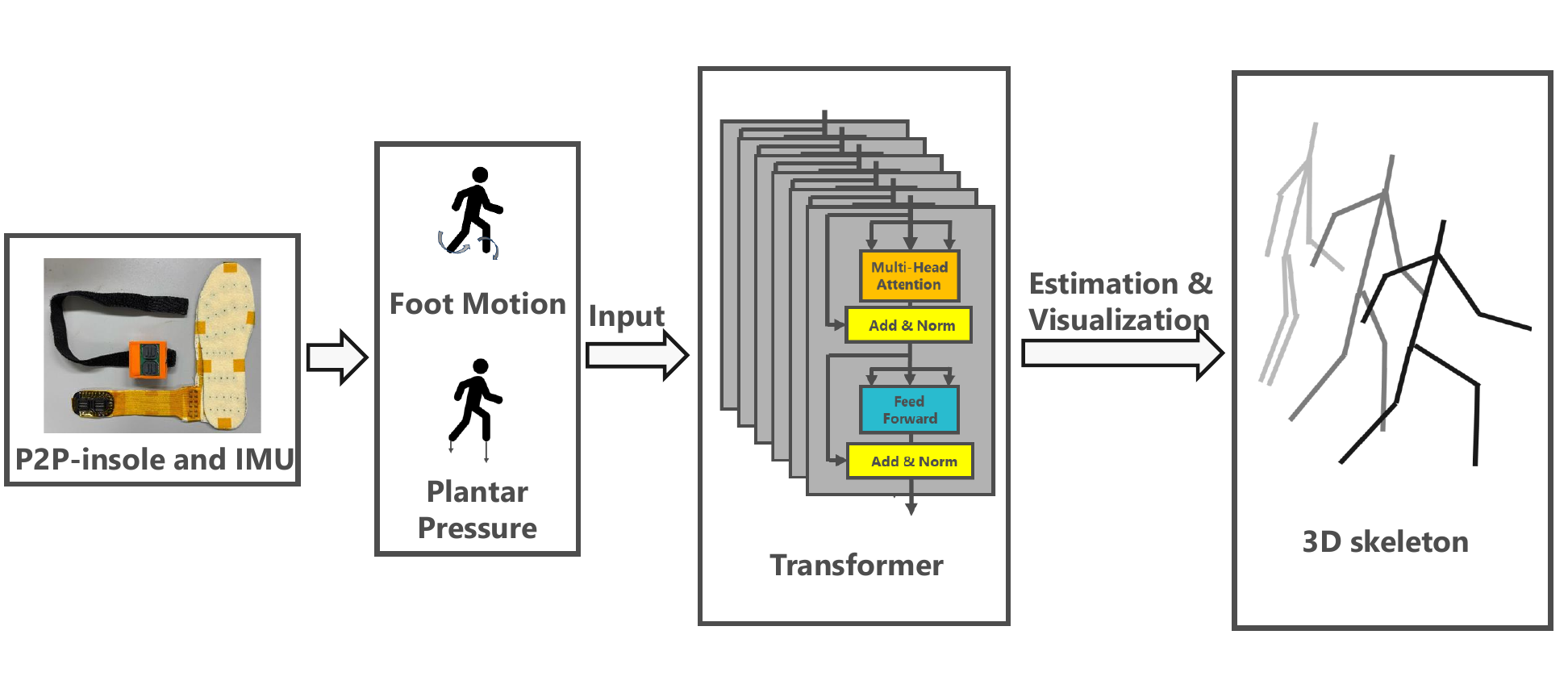}%
\end{wrapfigure}%
\begin{abstract}
This work presents P2P-Insole, a low-cost approach for estimating and visualizing 3D human skeletal data using insole-type sensors integrated with IMUs. Each insole, fabricated with e-textile garment techniques, costs under USD 1, making it significantly cheaper than commercial alternatives and ideal for large-scale production. Our approach uses foot pressure distribution, acceleration, and rotation data to overcome limitations, providing a lightweight, minimally intrusive, and privacy-aware solution. The system employs a Transformer model for efficient temporal feature extraction, enriched by first and second derivatives in the input stream. Including multimodal information, such as accelerometers and rotational measurements, improves the accuracy of complex motion pattern recognition. These facts are demonstrated experimentally, while error metrics show the robustness of the approach in various posture estimation tasks. This work could be the foundation for a low-cost, practical application in rehabilitation, injury prevention, and health monitoring while enabling further development through sensor optimization and expanded datasets.
\end{abstract}

\end{minipage}}}

\maketitle


\section{Introduction}
\label{sec:introduction}
\IEEEPARstart{H}{uman} posture is closely related to health and lifestyle; thus, the accuracy of evaluating human posture is important in rehabilitation and injury prevention. Although camera-based systems for posture estimation are effective in controlled settings, they have many drawbacks, including privacy, lighting, and occlusion. Advances in 2D and 3D pose estimation, including multi-camera setups and two-stage approaches \cite{cao2017realtime}\cite{iskakov2019learnable}\cite{reddy2021tessetrack}, have improved accuracy but remain impractical for everyday use due to environmental and computational constraints \cite{zhang2024mmvp}\cite{guo2022identity}\cite{wu2024soleposer}.
Body-worn sensors like IMUs offer flexibility but are prone to drift and discomfort during dynamic activities \cite{huang2018deep}. Hybrid methods combining RGB, LiDAR, and IMU data have shown promise \cite{furst2021hperl}\cite {pan2023fusing}, and contact-based datasets have improved accuracy by integrating force and pressure information \cite{hassan2021populating}. 

Pressure-sensing fabrics, such as those used in bedsheets, carpets, and clothing, enable applications like sleep posture classification \cite{davoodnia2021bed} \cite{lei2024pressure} and 3D skeleton estimation \cite{zhao20243d} \cite{wu2024soleposer}. However, these systems face challenges like spatial discontinuities, high costs, and limited sensing areas \cite{seong2024intelligent}.

In this study, we present P2P-Insole (Pressure to Posture Insole), a system that integrates machine learning techniques with insole-type sensors to estimate and visualize 3D skeletal data. The system is lightweight, minimally intrusive, and allows users to visually assess their posture, making it suitable for applications in rehabilitation support and injury prevention.

Our main contributions are as follows:
\begin{itemize}
\item Developed P2P-Insole (Motion Capture Insole), a low-cost embroidery-fabricated insole sensor with 35 pressure sensors, enabling detailed plantar pressure data collection for accurate 3D skeletal estimation.
\item Verified the relationship between sensor deployment and measurement errors, providing insights to optimize sensor placement for enhanced accuracy.
\item Introduced the first and second derivatives into the input stream of the Transformer model and verified their significance through controlled experiments, demonstrating improved model performance.

\end{itemize}

These contributions aim to enhance the accuracy of 3D skeletal estimation using plantar pressure data while ensuring privacy-conscious, user-friendly, and flexible applications in health monitoring and rehabilitation.


\section{Related Works}

\begin{table*}[ht]
\centering
\caption{Comparison of related studies and P2P-Insole (This Work)}
\label{tab:comparison}
\begin{tabular}{|l|l|l|l|l|l|}
\hline
\textbf{Study} & \textbf{Sensor Type} & \textbf{Auxiliary Data} & \textbf{Method} & \textbf{Accuracy} & \textbf{Application / Posture Target} \\ \hline
\textbf{P2P-Insole (ours)} & Insole (variable) & IMU  & Transformer  & \textless 75mm & Daily activities \\ \hline
SolePoser & Insole (32 points) & IMU & Two-stream Transformer & \textless 70mm (real-time) & Sports \& daily actions \\ \hline
PIFall & Insole (5x9 grid) & None & ResNet(2+1)D & Classification Accuracy: 91\% & Fall detection \\ \hline
Smart Chair & Seat pressure sensors & None & CNN & 93.13mm & Sitting posture \\ \hline
Intelligent Seat & Seat pressure & None & multi-class classification & Average error: 20.2cm & Sitting posture \\ \hline
\end{tabular}
\end{table*}

\subsection{Human pose Estimation}
Recent advancements in deep learning with large-scale datasets have raised the bar for 2D human pose estimation. Recovering 3D poses from 2D images is intrinsically challenging due to ambiguities and a lack of in-depth information. Although direct 3D recovery and two-stage approaches from 2D predictions, such as \cite{iskakov2019learnable}\cite{reddy2021tessetrack}, are promising, their performances are still limited. At the same time, temporal data and multi-camera setups have improved the system's accuracy, as in \cite{arunnehru2022machine}\cite{tang20233d}, occlusion and dependency on constrained scenarios remain important challenges.

Vision-based systems, including optical motion capture (MoCap) setups \cite{zhang2024mmvp}, provide precise tracking but are expensive and require extensive setups involving markers and infrared cameras \cite{guo2022identity}. Recent developments in deep neural networks have achieved single-camera 3D pose estimation, with most research exploring two-stage approaches for improved accuracy \cite{wu2024soleposer}. These methods are intensive in computational requirements and sensitive to occlusion, illumination, and privacy issues; hence, they are not very practical for day-to-day applications \cite{ye2022faster}. Accordingly, body-worn sensors like IMUs allow for more flexibility and natural tracking. Early methods like Sparse Inertial Poser and Deep Inertial Poser \cite{huang2018deep} used 6–17 IMUs. Despite these advances, IMU-based methods face issues such as drift and discomfort during dynamic activities. Hybrid approaches have emerged to address the limitations of vision- and sensor-based techniques, combining modalities like RGB, LiDAR, and IMU data to enhance pose estimation \cite{furst2021hperl}\cite{patil2020fusion}\cite{pan2023fusing}. Contact-based datasets have improved accuracy by incorporating pressure and force information \cite{hassan2021populating}. 

Our proposed system resolves these challenges by estimating 3D skeletal poses from pressure maps. Our contribution offers a robust alternative to the existing pose estimation methods, focusing on a practical and user-friendly design.

\subsection{Pressure-based Pose Estimation}
Pressure-sensing fabrics have been widely utilized to infer human poses by analyzing pressure distribution \cite{zhan2024satpose}. Pressure data from bedsheets \cite{davoodnia2021bed}\cite{lei2024pressure}, carpets \cite{luo2021intelligent}\cite{davoodnia2023human}, and clothing have been used for applications such as sleep posture classification \cite{seong2024intelligent}, 3D skeleton estimation \cite{zhao20243d}\cite{wu2024soleposer}\cite{wu2024dual} and motion capture related to fall detection \cite{guo2024pifall}. Despite their utility, such systems face challenges like spatial discontinuities in pressure patterns and limited sensing areas, complicating their adaptation to diverse environments. Wearable pressure sensors have also advanced human pose estimation \cite{zhang2024learn} implemented a deep learning pipeline to infer 3-D human poses using the full-body pressure data. Despite these efforts, most wearable systems face challenges like limited data availability, drift issues, and the inability to address full-body dynamics comprehensively.  

In summary, while previous works focus on limited actions or rely on vision-based systems, our method exclusively utilizes insole-type sensors to estimate 3D skeletal data. Our system eliminates cameras; hence, all privacy-related issues are removed. It operates independently of environmental factors such as lighting or occlusion and is minimally intrusive.


\section{System Design}
\begin{figure}[h]
  \centering
  \begin{minipage}{0.45\textwidth}
    \centering
    \includegraphics[width=\textwidth]{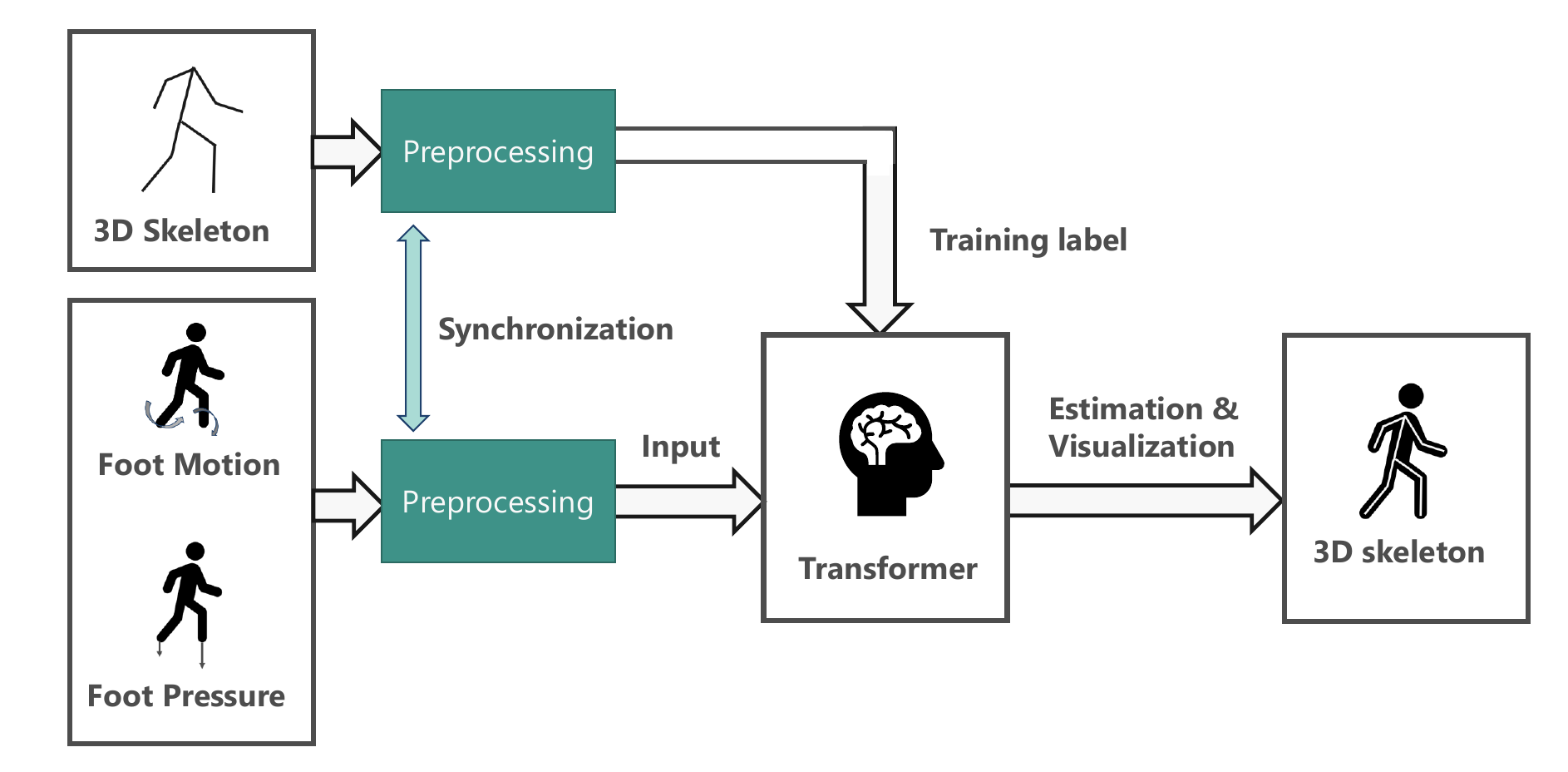}
    \caption{System Design}
    \label{fig:systemdesign}
    \end{minipage}
  \hfill
  \begin{minipage}{0.45\textwidth}
  \centering
  \includegraphics[width=\textwidth]{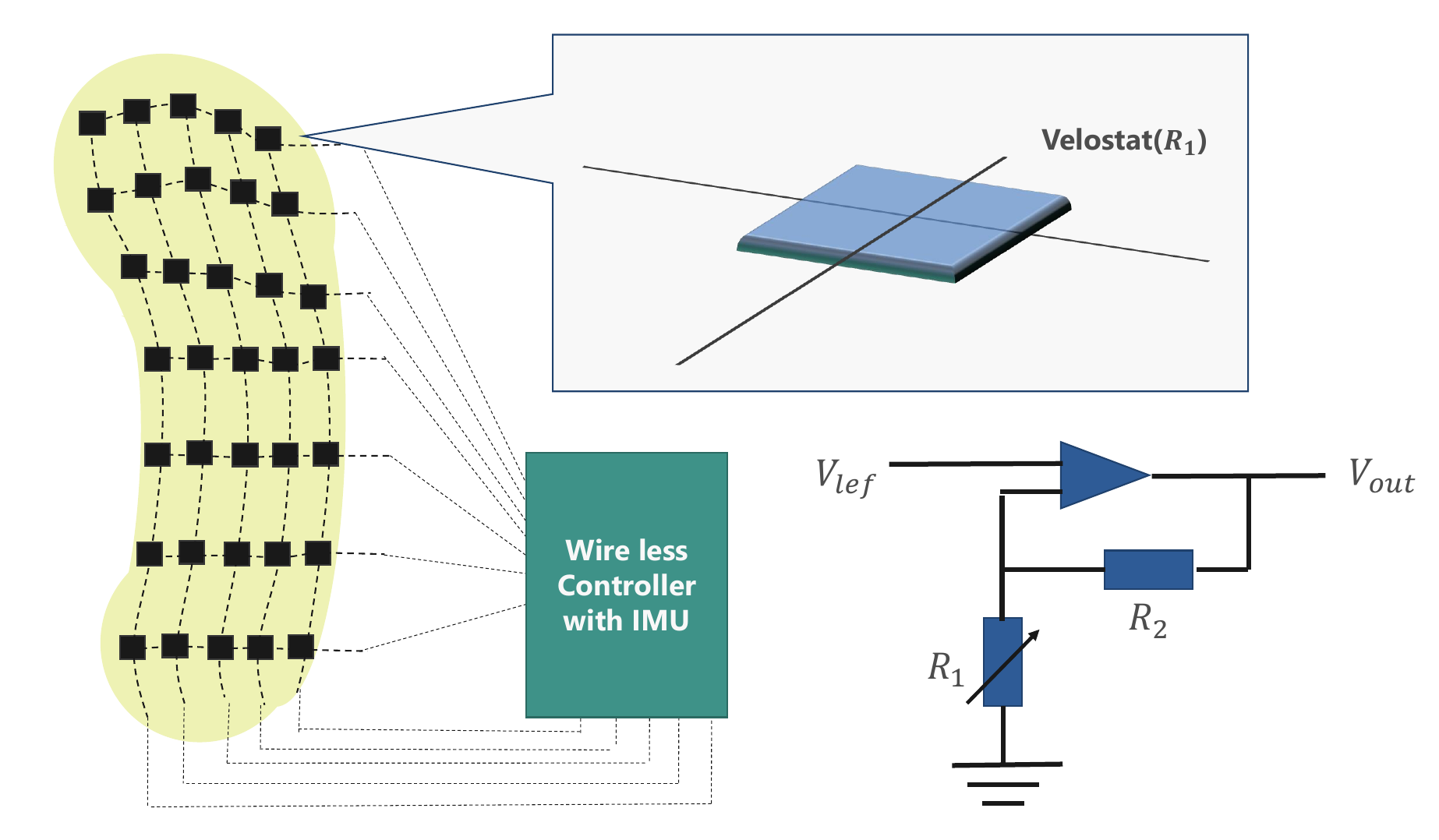}
  \caption{Insole Sensor Design and Non-Inverting Amplifier Circuit}
  \label{fig:insolesensor}
  \end{minipage}
\end{figure}

Figure \ref{fig:systemdesign} shows the system's architecture of the P2P-Insole. The system inputs foot pressure distribution, acceleration, and rotation information collected by an insole sensor and an inertial measurement unit (IMU). It uses a Transformer model to predict a 3D skeleton. The model is trained using a dataset that simultaneously captures data collected from the insole sensor and IMU, along with the corresponding 3D skeleton data using high precision Opti-Track mocap system. We developed a system to predict 3D skeletal structures integrating foot pressure, acceleration, and rotation data from an insole sensor and an IMU. The system employs a Transformer model trained on synchronously captured datasets, including insole sensor data, IMU data, and 3D skeletal data. The workflow consists of four steps: (1) simultaneous data collection across modalities, (2) preprocessing with normalization and feature enhancement to improve learning efficiency, (3) synchronization of all datasets based on timestamps, and (4) training the Transformer model with the preprocessed data for experimental validation.
The insole sensor, shown in figure \ref{fig:insolesensor}, features 35 pressure sensors utilizing a piezoresistive Velostat-based non-inverting amplification circuit. A low-cost ESP32 microcontroller ensures high-performance data acquisition at 100 Hz, while the custom circuit board integrates an IMU to collect pressure and motion data simultaneously.

\begin{equation}
V_{\text{out}} = \left(1 + \frac{R_2}{R_1}\right)V_{\text{lef}}
 \label{voltagediv}
\end{equation}

We implemented a voltage divider circuit on the PCB housing the ESP32 microcontroller. In this circuit, \(R_1\) and \(R_2\) form a voltage divider, with an input voltage \(V_{\text{lef}}\) applied to the circuit. The output voltage  \(V_{\text{out}}\) is determined by the voltage divider ( equation \ref{voltagediv}), which enables high-frequency measurements for efficient acquisition of pressure data with high temporal resolution.

\section{Method}
\begin{figure*}[t]
  \centering
  \includegraphics[scale=0.5]{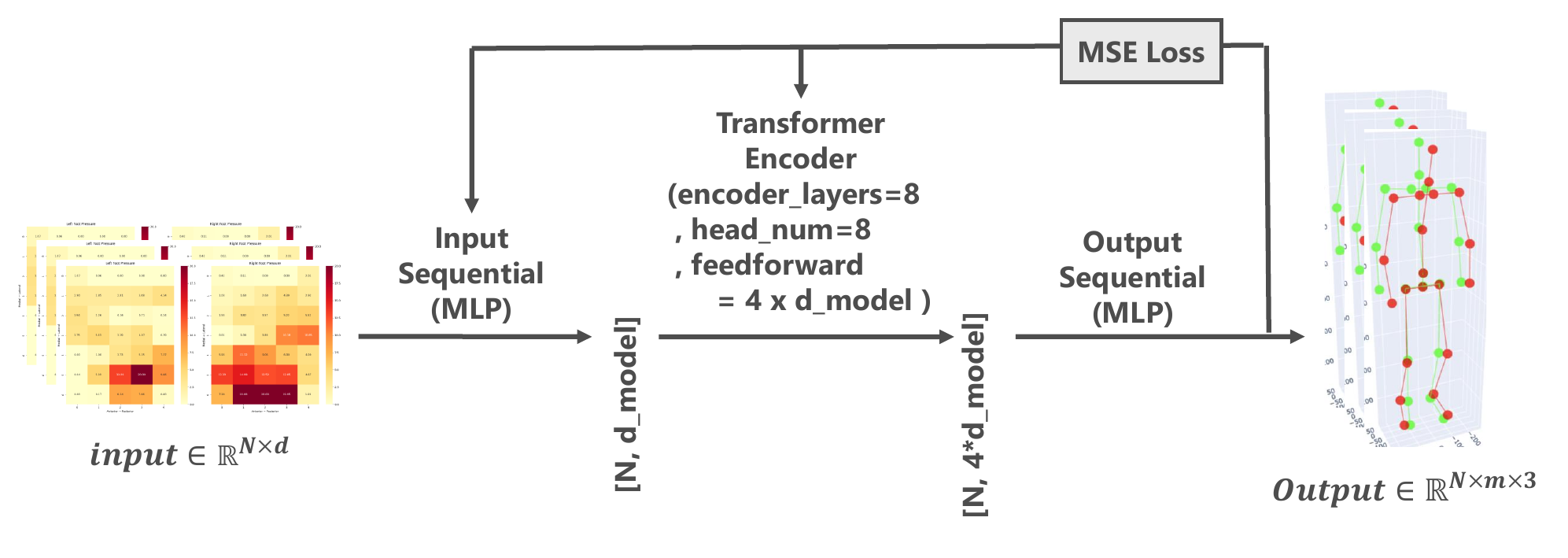}
  \caption{Transformer Architecture}
  \label{model}
\end{figure*}

We adopted a Transformer-based architecture for processing time-series data, as illustrated in figure \ref{model}. The core learning model was a 512-dimensional Transformer Encoder composed of 8 layers, each incorporating a multi-head self-attention mechanism with 8 heads. The input to the model, denoted as \( \mathbf{X} \), is a time-series dataset where each sample \( \mathbf{X} \in \mathbb{R}^d \), with \( d \) representing the dimension of the input features derived from the concatenation of pressure, rotation, and acceleration data, along with their respective derivatives. The entire input can be expressed as \( \mathbf{X} \in \mathbb{R}^{N \times d \times 2} \), where \( N \) is the number of samples (frames) in the dataset. The output layer was designed to predict 3D skeletal data, denoted as \( \mathbf{Y} \), where each sample \( \mathbf{Y} \in \mathbb{R}^{m \times 3} \), with \( m \) being the number of joints in the skeleton and 3 representing the \( x \), \( y \), and \( z \) coordinates of each joint. Consequently, the entire output can be expressed as \( \mathbf{Y} \in \mathbb{R}^{N \times m \times 3} \). To prevent overfitting, a dropout rate of 0.1 was applied throughout the architecture.

The input vector $\mathbf{X}$ consists of pressure distribution, rotational, and acceleration data:

\[
\mathbf{x} = (x_0, x_1, \dots, x_{d})
\]

\[
\begin{aligned}
    \text{Pressure sensor data} &: 35 \text{ dimensions} \\
    \text{3-axis rotational sensor data} &: 3 \text{ dimensions} \\
    \text{3-axis accelerometer data} &: 3 \text{ dimensions} \\
    \text{Total dimensions per foot} &: 35 + 3 + 3 = 41 \\
    \text{Total dimensions for both feet} &: 82
\end{aligned}
\]

The output $\mathbf{Y}$ represents the predicted 3D skeletal structure:

\[
\mathbf{y} = \{ j_0, j_1, \dots, j_{m-1} \}
\]

\[
\begin{aligned}
    \text{Number of joints} &: m = 21 \\
    \text{Each joint has 3 coordinates} &: (x, y, z) \\
    \text{Total output dimensions} &: 21 \times 3 = 63
\end{aligned}
\]

\begin{figure}[h]
  \centering
  \includegraphics[scale=0.2]{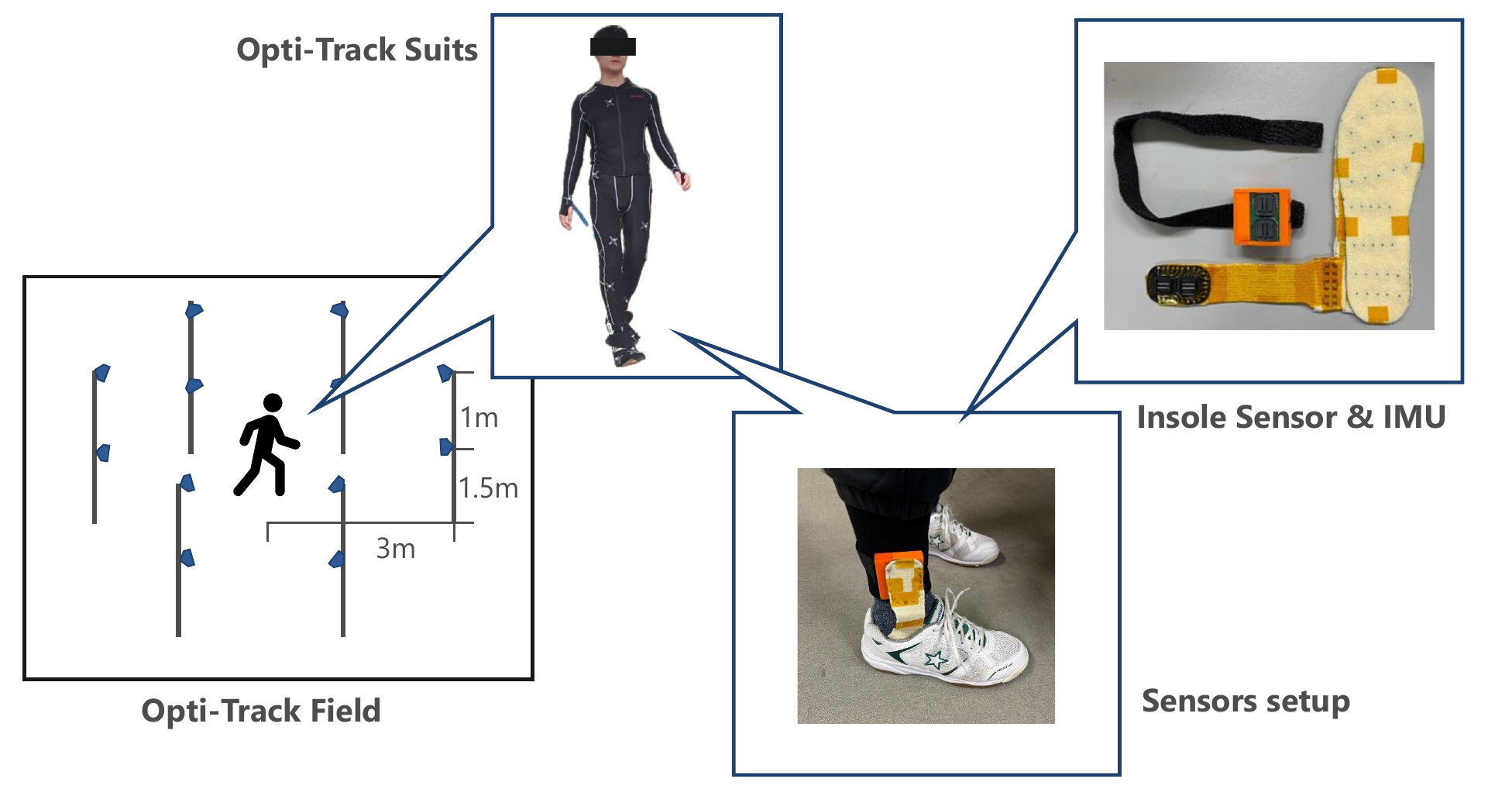}
  \caption{Experimental Setup}
  \label{ExperimentPictures}
\end{figure}

The training process utilized Mean Squared Error (MSE) as the loss function, optimized with the AdamW optimizer. The learning rate was initialized at 0.0005, with a weight decay of 0.001. Additionally, a ReduceLROnPlateau scheduler dynamically adjusted the learning rate when no improvement in validation loss was observed. The model was trained for 200 epochs with a batch size of 32. The dataset was split into training and validation sets in an 8:2 ratio, and the final model was saved when the validation loss reached its minimum.

\begin{equation}
\text{MSE Loss} = \frac{1}{N} \sum_{i=1}^{N} \left( y_{\text{pred}, i} - y_{\text{true}, i} \right)^2
\end{equation}

PyTorch served as the training framework, with training conducted on an NVIDIA GeForce RTX 4070 GPU. The selected hyperparameters and architectural components, including the Transformer’s multi-head self-attention mechanism and dropout regularization, were based on established practices for processing time-series data and validated through preliminary experiments.

\begin{figure*}[t]
  \centering
  \includegraphics[scale=0.5]{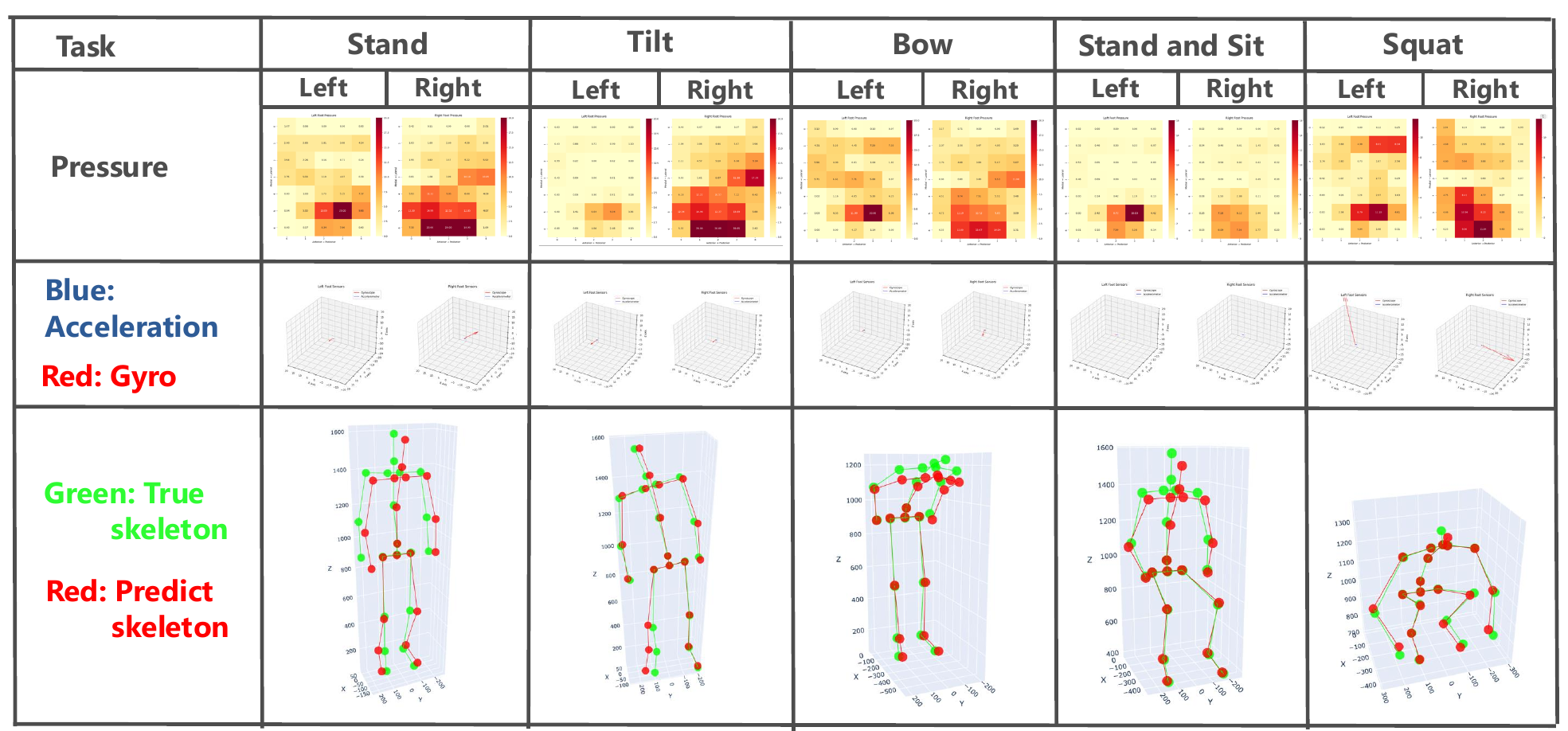}
  \caption{Highly accurate instantaneous predicted 3D skeleton.}
\end{figure*}

\section{Experiments and Results}

\subsection{Dataset}
The training data used in this study comprises three complementary datasets, each designed to enhance the 3D skeleton prediction model using a Transformer. The first dataset, collected with the OptiTrack motion capture system as shown in figure \ref{ExperimentPictures}, contains highly accurate 3D skeletal data recorded in a controlled studio environment with 12 Prime 13 cameras. Subjects wore OptiTrack suits, and the resulting dataset includes time-series 3D coordinates of 21 skeletal joint points, exported in CSV format as ground truth for model training.

The second dataset is on foot pressure distribution, which is captured using a low-cost insole-type pressure sensor developed in our laboratory. This sensor was fabricated through an embroidery-based method and offers significantly lower cost than commercial insoles with similar functionality while maintaining reliable performance. It features 35 pressure points arranged to conform to the foot's natural shape, recording time-series pressure values that provide detailed insights into plantar pressure patterns. The data were also exported in CSV format. The third dataset includes foot motion data from an IMU sensor attached to the subject's ankle. Foot motion data from the third dataset was obtained through an inertial measurement unit sensor attached to the subject's ankle. It offers time-series 3D vector acceleration and angular velocity data and supplements the pressure and skeletal data by incorporating dynamic foot motion. These datasets give a strong basis for training and validating the proposed model.

\subsection{Pose Label Generation}

\begin{figure}[h]
  \centering
  \includegraphics[scale=0.2]{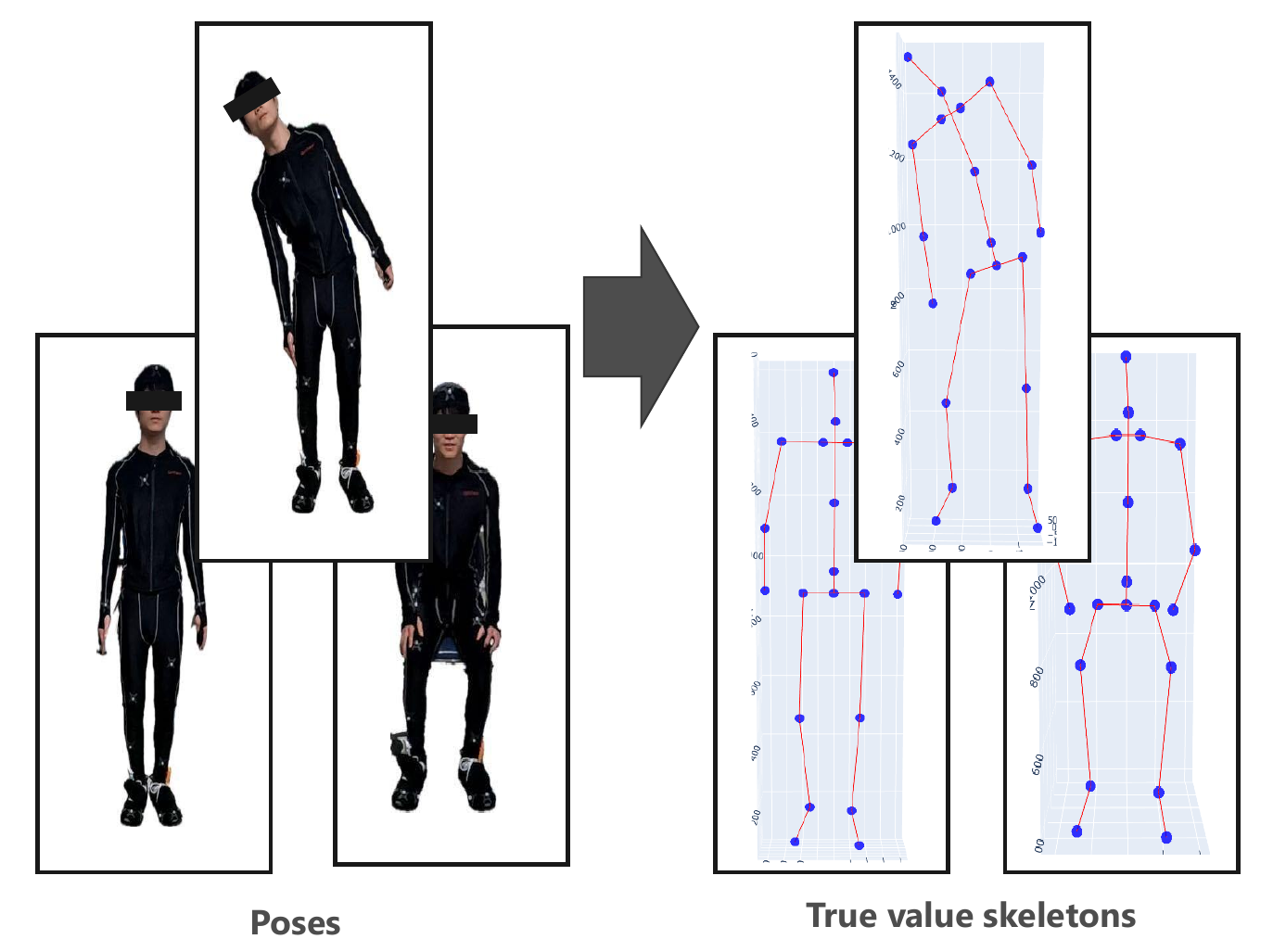}
  \caption{Pose Label Generation}
  \label{Label Generation}
\end{figure}

We designed and implemented a structured pipeline to collect data on eight fundamental movements commonly observed in daily life. The target movements as shown in figure \ref{Label Generation}, included tilting the body to the left and right, bowing, squatting, standing and sitting, standing on one leg, walking, jumping, and hopping on one leg. Four participants performed these movements in a controlled setting. Each participant repeated each movement continuously for two minutes, followed by six minutes of unrestricted free movement to capture natural variations. This resulted in a total data collection time of twenty minutes per participant. Across all four participants, we collected eighty minutes of synchronized movement data, providing a comprehensive dataset for further analysis.

\subsection{Preprocessing}
The preprocessing stage consisted of multiple steps to ensure data consistency and optimize model performance. For the 3D skeleton data, missing values were imputed using OptiTrack’s built-in data processing functionality. A low-pass filter was then applied to remove high-frequency noise, followed by downsampling to align timestamps at 0.01-second intervals. This ensured a uniform and noise-free dataset for skeletal joint data. The preprocessing of foot pressure distribution and foot motion data was performed jointly to maintain alignment. Missing values were filled with zero, and a moving average filter was applied to reduce noise. The combined data were then downsampled to match the 0.01-second intervals used for the 3D skeleton data. To facilitate efficient model training, normalization (eq. \ref{normalization}) and standardization (eq. \ref{standarization}) were applied across all data points. Additionally, feature enhancement techniques were introduced, where the first and second derivatives (eq. \ref{derivative}) of the time-series data were computed to capture detailed temporal changes.

\begin{equation}
x_{\text{norm}} = \frac{x - \min(x)}{\max(x) - \min(x)}
\label{normalization}
\end{equation}

\begin{equation}
x_{\text{std}} = \frac{x_{\text{norm}} - \mu}{\sigma}
\label{standarization}
\end{equation}

\begin{equation}
x' = \frac{\partial x_{\text{std}}}{\partial t}, \quad x'' = \frac{\partial^2 x_{\text{std}}}{\partial t^2}
\label{derivative}
\end{equation}

After preprocessing, the 3D skeleton, foot pressure, and motion data were synchronized based on their timestamps. This process ensured that all data sources were precisely aligned in terms of starting points, ending points, and the total number of frames. The final preprocessed dataset was prepared for training, providing a consistent and comprehensive input for the model.

\subsection{Performance Evaluation}
To evaluate the performance of the proposed system, we calculated the error between the predicted 3D skeleton data and the ground-truth 3D skeleton data using the Euclidean distance for each joint point. The primary evaluation metric was the Root Mean Square Error (RMSE) as in equation \ref{rmse}, computed across all frames, providing a quantitative measure of joint localization accuracy. Additionally, statistical metrics such as the mean and standard deviation of errors were analyzed to assess the model's performance for each joint. The mean RMSE across all joint points was used as an overall indicator of the system's accuracy. 

\begin{equation}
\text{RMSE} = \sqrt{\frac{1}{N} \sum_{i=1}^N (y_i - \hat{y}_i)^2}
\label{rmse}
\end{equation}

\begin{table}[t]
  \centering
  \caption{RMSE and Error Analysis Summary (mm)}
  \label{tbl:error_summary}
  \resizebox{\linewidth}{!}{%
  \begin{tabular}{|l|c|c|c|c|c|}
    \hline
    Task & Stand & Tilt & Bow & Stand and Sit & Squat \\
    \hline
    RMSE & 52.6 & 48.9 & 63.7 & 70.1 & 75.2 \\
    Median\_error & 37.4 & 34.6 & 47.9 & 47.4 & 46.1 \\
    Std. Dev. Error & 54.0 & 52.6 & 58.5 & 68.2 & 84.0 \\
    \hline
  \end{tabular}%
  }
\end{table}

\begin{table}[htbp]
\centering
\caption{RMSE and Error for Each Part of the Body Summary (mm)}
\begin{tabular}{lcccc}
\toprule
\textbf{Part} & \textbf{Task} & \textbf{Median Error} & \textbf{Std. Dev. Error} \\
\midrule
\multirow{4}{*}{Head} 
& Stand & 43.8 & 51.2 \\
& Tilt  & 42.4 & 53.7 \\
& Bow  & 71.1 & 66.2 \\
& Stand and Sit & 63.6 & 57.3 \\
& Squat & 60.1 & 67.2 \\
& Average & 56.2& 59.1\\

\midrule
\multirow{4}{*}{Spine} 
& Stand  & 4.6 & 18.2 \\
& Tilt  & 5.6 & 17.6 \\
& Bow  & 6.1 & 25.5 \\
& Stand and Sit  & 5.9 & 22.6 \\
& Squat  & 5.5 & 25.1 \\
& Average & 5.5& 21.8\\

\midrule
\multirow{4}{*}{Arms} 
& Stand & 45.0 & 45.1 \\
& Tilt & 43.8 & 47.5 \\
& Bow & 55.5 & 52.9 \\
& Stand and Sit & 55.7 & 48.5 \\
& Squat & 58.4 & 58.4 \\
& Average & 51.6& 50.4\\

\midrule
\multirow{4}{*}{Legs} 
& Stand & 38.0 & 56.7 \\
& Tilt & 32.9 & 49.0 \\
& Bow & 48.1 & 52.5 \\
& Stand and Sit & 53.3 & 80.9 \\
& Squat & 45.5 & 102.5 \\
& Average & 43.5& 68.3\\

\bottomrule
\end{tabular}
\end{table}

\begin{figure}[h]
  \centering 
  \includegraphics[scale=0.4]{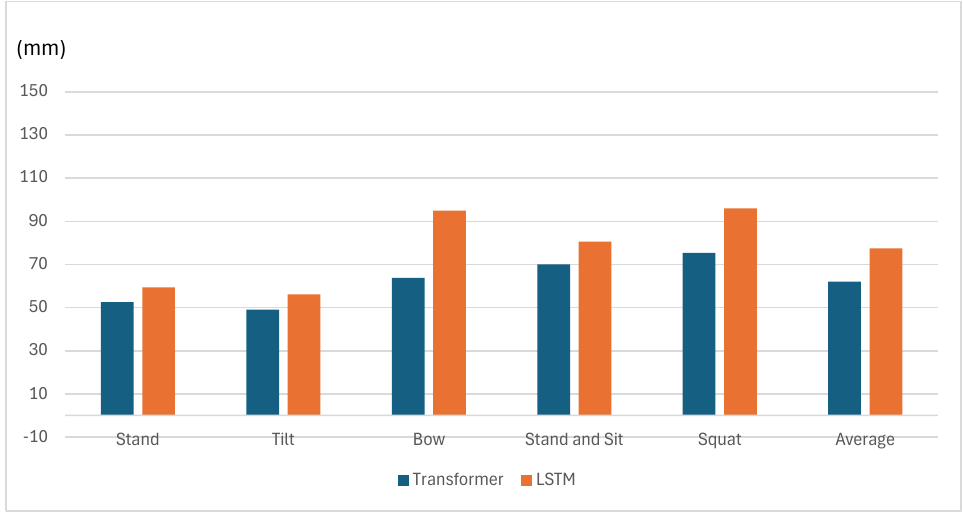}
  \caption{RMSE Comparison of Transformer and LSTM}
  \label{fig:comptransf}
\end{figure}

\begin{figure}[h]
  \centering
  \includegraphics[scale=0.4]{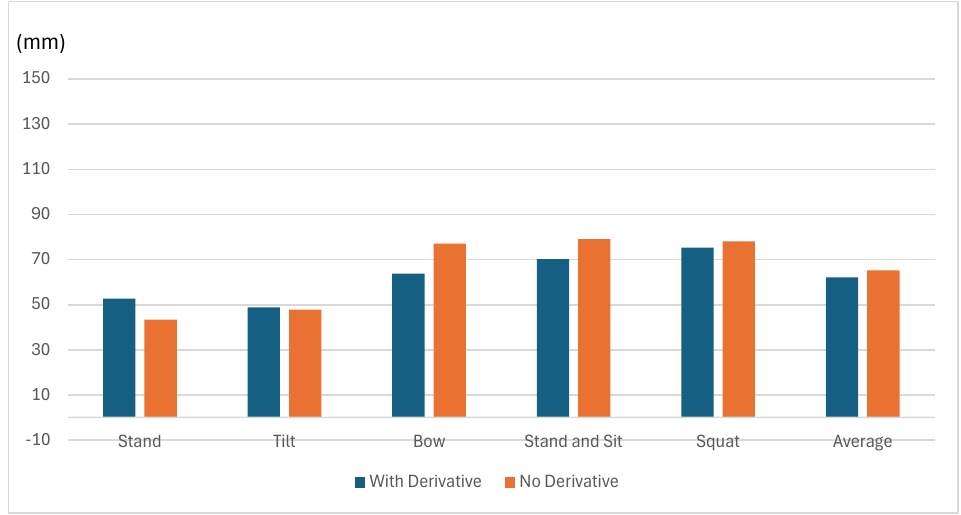}
  \caption{RMSE of comparison with and without derivative data}
  \label{fig:compdeff} 
\end{figure}

As shown in figure \ref{fig:comptransf}, the Transformer model's performance surpassed that of the LSTM model in all the tasks evaluated. The Transformer model significantly improved performance, particularly for complex motions like Bow and Squat. These results highlight the model's superior ability to process time series data. First-order and second-order derivatives were introduced as preprocessing steps to evaluate the impact of differentiated data. 
Figure \ref{fig:compdeff} shows differentiated data improved performance in tasks with more significant body movements, such as Bowing and Squatting, but reduced accuracy in tasks with minimal movements, like Standing and Tilting. These findings suggest that differentiation enhances sensitivity to motion features but may adversely affect accuracy in small-movement tasks, highlighting a trade-off between sensitivity and task-specific performance.

\section{Conclusions}

In conclusion, we propose the P2P-Insole, a low-cost embroidery-fabricated insole sensor with 35 pressure sensors for 3D posture estimation by fusing foot pressure distribution and motion data. This insole sensor, fabricated using low-cost e-textile garment techniques, is significantly cheaper and costs less than 1 US dollar per unit, making it significantly cheaper than commercial alternatives and an ideal candidate for large-scale production and practical application adoption. Given the above correlation between sensor position and measurement error, we suggested optimizing sensor placement to attain high accuracy at good cost efficiency. Besides that, first and second-order derivatives inserted into the input of the Transformer helped capture better temporal features, which improved the performance. Although the present system achieves these promising results, our future work includes refinements like reducing the number of sensors without sacrificing accuracy would ease the design and increase practicality for real applications. Expanding the dataset regarding the diversity of motions and participants would improve robustness and generalization. The presented research provides a good platform for developing cost-effective and user-friendly solutions for 3D skeletal estimation for rehabilitation, sports performance analysis, and health monitoring applications.

\section*{Acknowledgment}

This work was supported by the NEDO Intensive Support for Young Promising Researchers (Grant Number 21502121-0), Collaborative Research with Toyota Motor Corporation, and JKA and its promotion funds from KEIRIN RACE.

\bibliographystyle{ieeetr}
\bibliography{references}

\begin{IEEEbiography}[{\includegraphics[width=1in,height=1.25in,clip,keepaspectratio]{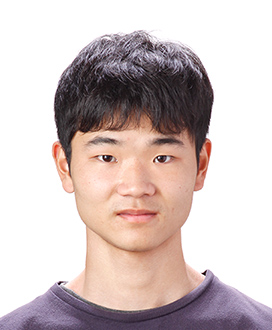}}]{Atsuya Watanabe} received his B.S. degree in Computer Science and Engineering from the University of Aizu, Japan, in 2025. He is working toward the M.S. degree in the Graduate School of Computer Science and Engineering, The University of Aizu, Japan. His research interests include human posture estimation and insole-type pressure sensors.
\end{IEEEbiography}

\begin{IEEEbiography}[{\includegraphics[width=1in,height=1.25in,clip,keepaspectratio]{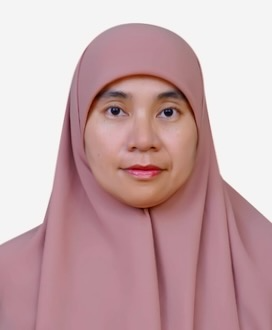}}]{Ratna Aisuwarya}  received the Master of Engineering degree from the Nara Institute of Science and Technology (NAIST), Japan, in 2012. She is currently pursuing the Ph.D. degree in the School of Computer Science and Engineering, University of Aizu, Japan. She is working as an Assistant Professor at the Department of Computer Engineering, Faculty of Information Technology, Andalas University, Padang, West Sumatra, Indonesia. Her current research interests include digital system design, Internet of Things (IoT), and wearable technology, with a focus on sensor development and integration in wearable systems.
\end{IEEEbiography}

\begin{IEEEbiography}[{\includegraphics[width=1in,height=1.25in,clip,keepaspectratio]{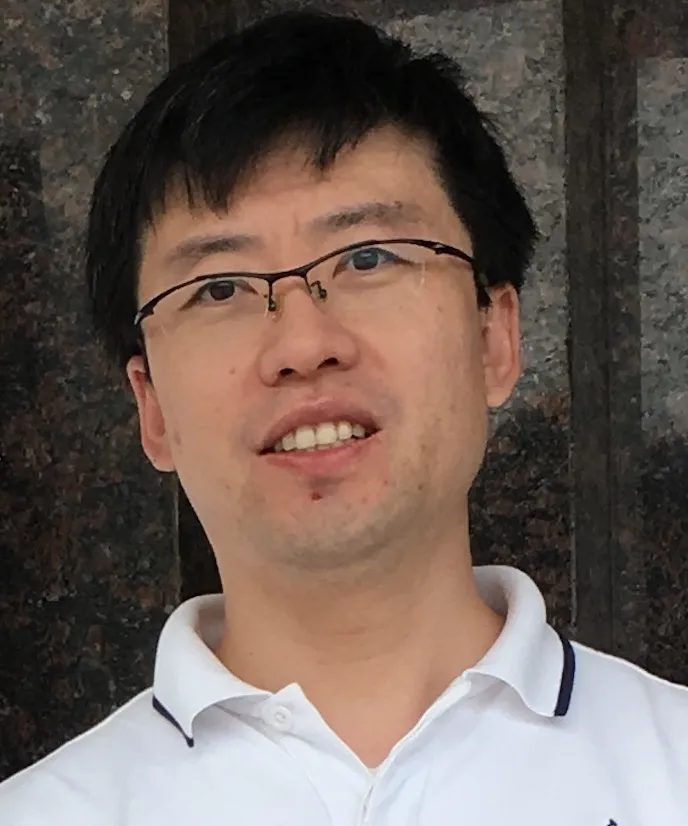}}]{Lei Jing} (M'12) received his Ph.D. degree in computer science and engineering from the University of Aizu, Japan, in 2008. He is currently a Senior Associate Professor at the School of Computer Science and Engineering, University of Aizu. His research interests include human position, posture, and motion tracking, soft circuit design, and the tactile internet. The applications of his work encompass human activity abnormality detection, sign language recognition, and human-robot interaction. He has published over 120 papers and holds six patents in related areas.
\end{IEEEbiography}

\end{document}